\documentclass[10pt,twocolumn,letterpaper]{article}

\usepackage[pagenumbers]{wacv} 

\usepackage{graphicx}
\usepackage{amsmath}
\usepackage{amssymb}
\usepackage{booktabs}
\usepackage{appendix}
\usepackage{mathrsfs}

\usepackage{enumitem}
\setlist[enumerate]{noitemsep, topsep=0pt}

\usepackage[pagebackref,breaklinks,colorlinks]{hyperref}

\usepackage[capitalize]{cleveref}
\crefname{section}{Sec.}{Secs.}
\Crefname{section}{Section}{Sections}
\Crefname{table}{Table}{Tables}
\crefname{table}{Tab.}{Tabs.}

\def\wacvPaperID{731} 
\def\confName{WACV}
\def\confYear{2024}

\begin{document}

\title{FIRE: Food Image to REcipe generation}


\author{Prateek Chhikara$^{1,2}$, Dhiraj Chaurasia$^{1,2}$, Yifan Jiang$^{1,2}$, Omkar Masur$^{2}$, and Filip Ilievski$^{1,2}$ \\
        $^{1}$Information Sciences Institute, Marina del Rey, USA \\
        $^{2}$University of Southern California, Los Angeles, USA \\
       \small \texttt{\{pchhikar,chaurasi,yifjia,ilievski\}@isi.edu} ~~~ \texttt{\{omasur\}@usc.edu}
}

\maketitle

\begin{abstract}
Food computing has emerged as a prominent multidisciplinary field of research in recent years. An ambitious goal of food computing is to develop end-to-end intelligent systems capable of autonomously producing recipe information for a food image. Current image-to-recipe methods are retrieval-based and their success depends heavily on the dataset size and diversity, as well as the quality of learned embeddings. Meanwhile, the emergence of powerful attention-based vision and language models presents a promising avenue for accurate and generalizable recipe generation, which has yet to be extensively explored. This paper proposes \texttt{FIRE}, a novel multimodal methodology tailored to recipe generation in the food computing domain, which generates the food title, ingredients, and cooking instructions based on input food images. \texttt{FIRE}
leverages the BLIP model to generate titles, utilizes a Vision Transformer with a decoder for ingredient extraction, and employs the T5 model to generate recipes incorporating titles and ingredients as inputs. We showcase two practical applications that can benefit from integrating \texttt{FIRE} with large language model prompting: recipe customization to fit recipes to user preferences and recipe-to-code transformation to enable automated cooking processes. Our experimental findings validate the efficacy of our proposed approach, underscoring its potential for future advancements and widespread adoption in food computing. The code is available at \url{https://github.com/prateekchhikara/FIRE}
\end{abstract}

\section{Introduction}
Food is not only a vital source of sustenance but also an integral part of our cultural identity, defining our lifestyle, traditions, and social interactions \cite{10.1145/3329168}. As the well-known saying goes, \textit{``Tell me what you eat, and I will tell you who you are,"} \cite{intro1} emphasizing the idea that an individual's dietary choices reflect their identity \cite{min2017you}. 
Moreover, a person's physical appearance and cognitive abilities often bear evidence of their dietary habits, as the selection of nutritious food contributes to the overall well-being of both the body and mind \cite{intro2}. The advent of social media enables anyone to share captivating visuals of personal experiences related to the delectable food they consume. A simple search for hashtags like \texttt{\#food} or \texttt{\#foodie} yields millions of posts, underscoring the immense value of food in our society \cite{social_media}.
The significance of food accompanied by its large amounts of publicly available data has inspired food computing tasks~\cite{10.1145/3329168} that associate visual depictions of dishes with symbolic information. An ambitious goal of food computing is to produce the recipe for a given food image, with applications such as food recommendation according to user preferences, recipe customization to accommodate cultural or religious factors, and automating cooking execution for higher efficiency and precision \cite{papadopoulos2022learningprogramrepresentations}.

\begin{figure}[!t]
	\centering
\includegraphics[width=\linewidth]{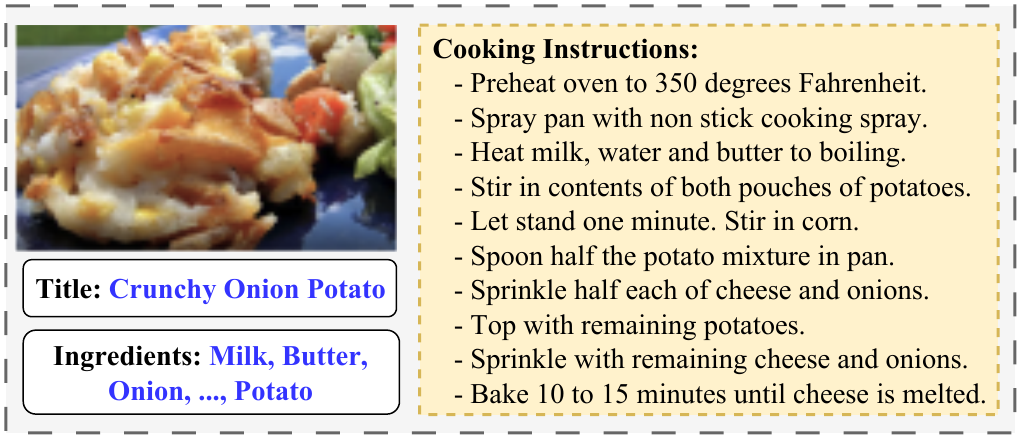}
	\caption{Given a potentially unseen image, our method \texttt{\textbf{FIRE}} generates a corresponding recipe consisting of a title, ingredients, and cooking instructions.
 }
	\label{main}
\end{figure}

Generating detailed recipe information or cooking procedures solely from a food image presents a considerable challenge \cite{salvador2019inverse,food_recipe_recommendation}. Food computing has been of interest to the computer vision (CV) community, whose efforts to use image processing for food quality assurance can be traced back to 1996 \cite{food_quality_1996}. State-of-the-art food image processing methods~\cite{kawano2014food,martinel2018wide,salvador2019inverse} use deep learning techniques to extract ingredients from images with limited success. Meanwhile, a popular natural language processing (NLP) application has been recipe generation, a procedural task of creating recipes based on a flexible set of ingredients as inputs. Typical models for recipe generation include~\cite{chef-transformer,salvador2019inverse,DBLP:journals/corr/abs-2110-01209}. We note that prior work has not connected the dots between the CV and NLP research in order to provide an end-to-end system that generates recipes from images. Moreover, current methods for food computing have not caught up with the most recent advances in NLP and CV, featuring diffusion models and language modeling.

This paper presents a novel multimodal methodology that we call \texttt{\textbf{FIRE}} (\textbf{F}ood \textbf{I}mage to \textbf{RE}cipe generation). \texttt{\textbf{FIRE}} is designed to generate comprehensive recipe output for the food computing domain, 
including food titles, ingredients, and cooking instructions, based on input food images as shown in Figure \ref{main}. We leverage recent advancements in CV and language modeling to employ state-of-the-art techniques that have demonstrated exceptional performance in various vision and language tasks. \texttt{\textbf{FIRE}} connects the dots between state-of-the-art (SotA) models, using BLIP \cite{blip1} for title generation, a Vision Transformer \cite{dosovitskiy2020image} with a decoder for ingredient extraction, and the T5 \cite{t5} model for cooking instruction generation. Furthermore, we highlight two practical applications that can benefit from integrating \texttt{\textbf{FIRE}} with prompting large language models: \textit{recipe customization} for personalized recipe adaptation, and \textit{recipe-to-code generation}, enabling automated cooking processes. The contributions of the paper are as follows: 

\begin{enumerate}
\itemsep0em
    \item We leverage the capabilities of Vision Transformers (ViT) \cite{dosovitskiy2020image} to get expressive embeddings from food images, which are subsequently fed into an attention-based decoder  to extract the ingredients of the recipe. 
    \item  We present an end-to-end pipeline for generating recipe titles and cooking instructions, utilizing SotA vision (BLIP) and language (T5) models, respectively.
    \item Our multimodal approach outperforms the existing work based on two evaluation metrics: (a) set metrics for ingredient extraction and (b) document-level metrics for cooking instruction generation.
    \item We showcase the ability of \texttt{\textbf{FIRE}} to support two novel food computing applications: \textit{Recipe Customization} and \textit{Recipe to Code Generation}, through integration with few-shot prompting of large LMs.
\end{enumerate}
We organize this paper as follows; Section \ref{sec:related_work} of the paper gives a detailed overview of the related work in the field of food computing and its gap against SotA models. We describe our proposed methodology (\texttt{\textbf{FIRE}}) in Section \ref{sec:proposed_work}. Section \ref{sec:experiment_setup} describes the experimental setup we follow to obtain the results, which are presented in Section \ref{sec:results}. Section \ref{sec:applications} illustrates two advanced applications that can benefit from our proposed approach. Finally, we conclude our paper in Section \ref{sec:conclusion} with future research directions. We make all of our code available to stimulate work on recipe generation.

\section{Related Work} \label{sec:related_work}

\subsection{Food Computing}
Recently, the importance of food and the availability of extensive multimodal food datasets, such as Food-101 \cite{bossard2014food}, Recipe1M \cite{recipe1m}, and Recipes242k \cite{rokicki2018impact}, have enabled computational research on food computing tasks \cite{10.1145/3329168}.
We review prior work on food recognition and recipe generation.

\noindent \textbf{Food Recognition} is an image-to-text task requiring models to detect food categories in a food image. Recognition of food items can offer people comprehensive information and a better understanding of unfamiliar dishes, thereby improving other food-related applications as well \cite{pouladzadeh2017mobile, aguilar2018grab}. 
Previous works focus on extracting deep representations of food \cite{kawano2014food, 10.1145/2964284.2967205, martinel2018wide, salvador2019inverse}.
Martinel \textit{et al.} \cite{martinel2018wide} adapted a slice convolution block in the residual network to capture features in images. Salvador \textit{et al.} \cite{salvador2019inverse} proposed InverseCooking, an encoder-decoder framework to output the title of the food. 
Wang \textit{et al.} \cite{wang2020structure} also utilized images to get the recipe by treating it as an image captioning task.
Notably, earlier architectures are constrained as they tend to emphasize global features rather than local features and can not detect the ingredients overlapping in the image~\cite{10144465}. Instead, we employ a SotA vision encoder, ViT \cite{dosovitskiy2020image}, to enhance the extraction of the local semantic segmentation.


\noindent \textbf{Recipe Generation} is a more complex text-to-sequence task generating food recipes based on the ingredients provided. To solve this task, models must possess knowledge of food composition, ingredients, and cooking procedures to perform the task accurately. Early attempts at recipe generation were constrained by limited model capacity and structure, leading to solutions that relied on information retrieval techniques \cite{wang2008substructure, 5693849}. Wang \textit{et al.} \cite{wang2008substructure} developed a novel similarity and filtering algorithm to increase the search accuracy. Xie \textit{et al.} \cite{5693849}, leveraged the cooking flow and eating features with other domain knowledge  to enhance the searching process. More recent work relies on encoder-decoder structures to generate recipes~\cite{salvador2019inverse,DBLP:journals/corr/abs-2110-01209} with multimodal settings. Salvador \textit{et al.} \cite{salvador2019inverse} presented a framework that utilizes encoded image and ingredients representations in recipe generation. Wang \textit{et al.} \cite{DBLP:journals/corr/abs-2110-01209} added tree structures within the encoder-decoder process to incorporate structure-level information. 
In contrast to prior unimodal work, our approach uses images as input and generates titles and ingredients as an intermediate representation, and uses them to generate recipes. 

\begin{figure*}[!t]
	\centering
	\includegraphics[width=\linewidth]{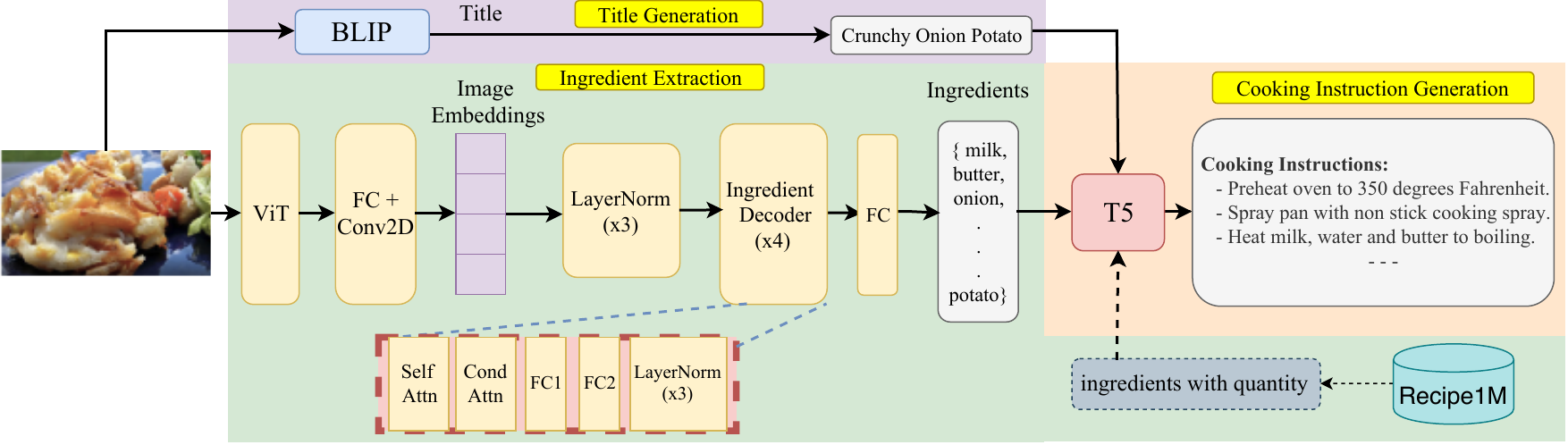}
	\caption{Proposed architecture to extract ingredients, and generate the recipe title and cooking instructions from a food image. (\textit{Ingredients with quantity is passed during the train time only})}
	\label{img:architecture}
\end{figure*}

\subsection{State-of-the-Art Models}
We review state-of-the-art models that have not found a broad application in food computing tasks to date.

\noindent \textbf{Image to Text Models} gradually play an important role in vision language tasks with the development of deep learning~\cite{krizhevsky2017imagenet}. Models follow a main pipeline that encodes image input into intermediate stages and then decodes them into text output~\cite{vinyals2015show,xu2015show}. Vinyals \textit{et al.}~\cite{vinyals2015show} encoded the input image into a global visual vector through CNN and then applied RNN to generate captions, whose generalization on multi-task also supported the effect of the encoder-decoder pipeline. Xu \textit{et al.} ~\cite{xu2015show} further built an attention mechanism to pick the most related subregion vectors rather than depending only on the global visual vector. Despite the progress in the encoder-decoder pipelines, their ability is limited by only emphasis on single modularity input. Recent research switched to unimodality with the birth of large image-text datasets~\cite{jia2021scaling,yao2021filip}. CLIP~\cite{radford2021learning} modified GPT-2~\cite{radford2019language} to obtain text features from textual input and used image-text contrastive learning, which trained the model with the similarity between the image and text. ALBEF~\cite{li2021align} utilized ViT (pre-trained on large amounts of data and transferred to multiple mid-sized or small image recognition benchmarks)~\cite{dosovitskiy2020image} as an image encoder and BERT~\cite{devlin2018bert} as a text encoder to extract information with attention. One additional multimodal
encoder was built on the extraction output, which additionally incorporated masked-language-modeling loss (MLM) to enhance the image-text interactions. Based on previous work, BLIP model \cite{blip1}, designed to achieve unified vision-language understanding and generation, 
reached state-of-the-art results on various vision-language tasks in a zero-shot manner \cite{zhang2023using}. 
By employing a captioner to generate synthetic captions and applying a filtering mechanism, BLIP maximizes the utilization of noisy web data.
In our work, we leverage BLIP and ViT to generate food titles and ingredients separately.

\noindent \textbf{Text (Sequence) to Sequence Models} takes a sequence of text as input and maps it into a succession of another sequence~\cite{yousuf2021systematic}. For this task, previous work either built recurrent neural networks (RNNs)~\cite{sutskever2014sequence,cho2014learning,bahdanau2014neural} or attention-based models~\cite{devlin2018bert,radford2019language,t5,vaswani2017attention}. GPT-2~\cite{radford2019language} is based on a transformer-decoder to perform tasks on various fields in a zero-shot setting, while T5 (text-to-text encoder-decoder model)~\cite{t5} transformed text-related tasks in a text-to-sequence format to enhance its general ability. With the recent progress of large pre-trained LMs, prompting has become a popular and efficient approach to tackle many NLP tasks \cite{liu2023pre}. More specifically, few-shot prompting provided input-output mapping in demonstration to guide LMs to prompt the specific structure. Chain-of-thought (CoT) \cite{weichain} with Self-Consistency \cite{wang2022self} reached state-of-art on commonsense reasoning and symbolic reasoning, even compared to supervised models.
Madaan \textit{et al.} \cite{madaan22emnlp} used graph demonstration to hint LMs generate complex Python classes for reasoning and state tracking. 
Capitalizing on these advancements, we employ T5 to generate cooking instructions from food ingredients and titles. To overcome resource and time constraints posed by large-scale food dataset, we also show how few-shot prompting can support food computing applications.
Our work analyzes how to generate a curated recipe or convert recipe into a structured program flow \cite{papadopoulos2022learningprogramrepresentations} for further application with the help of prompting.

\section{Proposed Methodology (\texttt{FIRE})}
\label{sec:proposed_work}
\textbf{FIRE} consists of three components:
\texttt{(1)} title generation from food images by using state-of-the-art image captioning, \texttt{(2)} ingredient extraction from images using vision transformers and decoder layers with attention, and \texttt{(3)} cooking instruction generation based on the generated title and extracted ingredients using an encoder-decoder model. 

\subsection{Title Generation}
We generate recipe titles from food images using the BLIP model, a state-of-the-art image captioning approach. In our initial experiments with the off-the-shelf BLIP model, we observed promising results, yet, BLIP's prediction accuracy was lower because of the domain shift between its training data and the food domain. Namely, BLIP tends to capture extraneous details impertinent to our goal because it was originally designed to provide a comprehensive image caption for a wide variety of settings. As an illustration, when presented with an image of a muffin, BLIP produced the description \textit{`a muffin positioned atop a wooden cutting board'}.
To better align the generated captions with recipe titles, we fine-tune the BLIP model using a subset of the Recipe1M dataset. We restrict our tuning to 10\% of the training dataset, as fine-tuning on the entire dataset is computationally intensive \cite{kim2023exposing}. We observe that the fine-tuned version of BLIP shows promising improvements in generating accurate, aligned, and pertinent titles for food images. The fine-tuned BLIP captions the same example image with a shorter string \textit{`muffin'}, removing the additional extraneous information. 

\subsection{Ingredient Extraction}
\label{subsec:title_ingredient}
Extracting ingredients from a given food image presents challenges due to the inherent complexity and variability of food compositions. Unlike generating titles or captions, determining a comprehensive and accurate list of ingredients requires a deep understanding of food characteristics, textures, and interactions. Additionally, the visual appearance of certain ingredients may overlap, whereas others may not be visible at all, leading to potential ambiguities and difficulties in discerning specific components. For instance, in a food image that contains a dish with melted cheese on top, from visual appearance alone it may be challenging to determine if the cheese used is mozzarella, cheddar, or any other type. These nuances motivate need for an architecture capable of retaining expressive embeddings from food images. 
While finetuned BLIP successfully generated accurate titles for food images, our analysis showed that using it for ingredient extraction led to significant hallucinations in the output. As BLIP is primarily trained for image captions, it struggles to generate ingredient lists accurately.
To address this challenge, we develop an ingredient extraction pipeline (shown in Figure \ref{img:architecture}) built on top of the one proposed by \cite{salvador2019inverse}. 

\noindent \textbf{Feature Extractor:} We extract the image's features by employing a vision transformer (ViT). ViT's attention mechanism enables for effective handling of feature representations with stable and notably high resolution. This capability precisely meets the requirements of dense prediction tasks such as ingredient extraction from food images \cite{zuo2022vision}.  Furthermore, transformer-based approaches exhibit minimal reliance on the inductive bias, facilitating effective interaction and integration of long-range information. 
Unlike conventional CNNs, the output of a ViT is sequential; therefore, we use a fully connected (FC) linear layer to reshape the output and pass it to a 2D convolution (Conv2D) layer. 

\noindent \textbf{Ingredient Decoder:} The feature extractor produces image embeddings. We pass these image embeddings through three normalization layers (\textit{layerNorm}) and subsequently feed the output into our ingredient decoder responsible for extracting ingredients. 
The decoder consists of four consecutive blocks, each comprising multiple sequential layers: self-attention, conditional attention, two fully connected layers, and three normalization layers. In the last step, the decoder output is processed by a fully connected layer with a node count equivalent to the vocabulary size, resulting in a predicted set of embeddings.

Given a corpus with ingredients and recipes corresponding to food images,
we construct a dictionary $D$ consisting of $N$ possible ingredients. Each recipe, $r_i$, is associated with a set $S$, comprising $K$ ingredients selected from this dictionary.
Given that the order of the ingredients does not affect the resulting recipe, we represent the ingredients as a set rather than a list. In other words, we exploit co-dependencies among ingredients without penalizing for prediction order. 
We represent the ingredient set $S$ using a binary vector, $s$, of dimension $N$, where $s_i$ = 1 if $s_i\in S$, and 0 otherwise. Consequently, our training dataset consists of $m$ pairs of image and ingredients sets: $\{(x_i , s_i )\}_{i=0}^{m}$. In this case, the goal is to predict $\hat{s}$ from an image \textit{x} by maximizing the following objective:
\begin{equation}
\arg\max_{\theta_{img},\theta_{ing}} \sum \log p(\hat{s}_i = s_i|x_i;\theta_{img},\theta_{ing})
\end{equation}
where $\theta_{img}$ and $\theta_{ing}$ represent the learnable parameters for the image encoder and ingredient decoder, respectively. While there may exist certain dependencies among the ingredients, such as the common combination of salt and pepper, these dependencies do not exert a dominant influence. Consequently, we can reasonably assume independence between the ingredients and factorize them as follows:
\begin{equation}
\sum\limits_{j=1}^{N} \log p(\hat{s}_{ij} = s_{ij}|x_i) \leftarrow p(\hat{s}_i = s_i|x_i) 
\end{equation}
Our decoder makes ingredient predictions sequentially until it encounters an end-of-sequence (EOS) token. To mitigate the impact of the order of the ingredients, we aggregate the outputs separately across different time steps and use max pooling at the end to obtain the ingredient set. This enables training the model using binary cross-entropy loss ($loss_{ingr}$) between the predicted ingredients (after pooling) and the ground truth. However, since the EOS information is lost during pooling, we use a custom EOS loss ($loss_{eos}$). This loss calculates the binary cross-entropy between the predicted EOS probabilities at all time steps and the corresponding ground truth. Furthermore, to enhance performance, we incorporate a cardinality L1 penalty ($loss_{card}$), which constrains the length of the predicted ingredients to be close to the ground truth ingredients. We  empirically find that integrating the $loss_{card}$ leads to better performance. 
\begin{equation}
    loss = \alpha_1\times~loss_{ingr} + \alpha_2\times~loss_{eos} + \alpha_3\times~loss_{card}
\end{equation}
where, $\alpha_1$=100 , $\alpha_2$=1 , and $\alpha_3$=1 are the hyper-parameters.

\begin{figure}[!h]
\centering
\includegraphics[width=\linewidth]{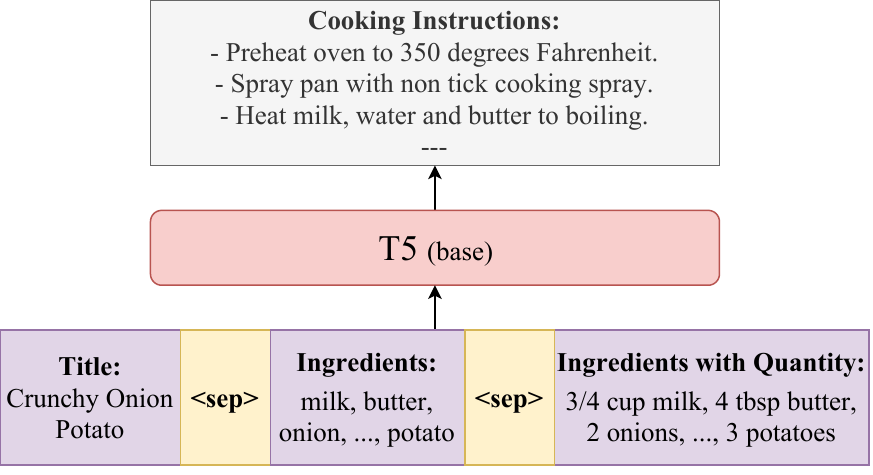}
	\caption{Generating cooking instructions for a title and a set of ingredients. (ingredients with quantity is present only during the fine-tuning of T5)}
	\label{t5}
\end{figure}

\subsection{Cooking Instruction Generation} \label{subsec:recipe_generation}
Considering the remarkable accomplishments of LMs in natural language applications like text generation and question answering \cite{bionlp}, we pose cooking instruction generation as a language modeling task. 
Large LMs such as GPT \cite{brown2020language}, LLaMa \cite{touvron2023llama}, and Alpaca \cite{alpaca} are pre-trained with billions of tokens with multiple training objectives, which makes them capable of understanding language in context. Refining the LMs for downstream tasks has demonstrated remarkable outcomes in various NLP assessments. While we expect that large LMs would be capable of generating cooking instructions after fine-tuning, they require prohibitive computational resources given their large number of parameters. Given the available resources and our research objective, we adopt popular encoder-decoder model, T5 \cite{t5}, for generating cooking instructions.
We conduct all experiments using base T5 model with 220M parameters. During finetuning, we pass title and ingredients of the recipe as a formatted string (see Figure \ref{t5}), inspired by prior work~\cite{zhou2021detecting}. 

The T5 is finetuned on three inputs: title, ingredients, and ingredients with quantity to incorporate maximum information from the dataset. However, we do not have ingredients with quantity at inference time; hence we can pass only the title and ingredients. Moreover, excluding the quantity information from our model ensures a fair comparison with previous approaches and investigates whether our model's advantage stems from a well-structured architecture rather than relying solely on the augmentation of additional knowledge. By removing the influence of quantity information during inference, we aim to highlight the inherent capabilities of T5 and its ability to generate high-quality cooking instructions. 


\section{Experiment Setup}
\label{sec:experiment_setup}

\subsection{Dataset}
Recipe1M is a large-scale dataset of over one million cooking recipes
\cite{recipe1m}. The dataset contains rich food-related information, including recipe titles, ingredients, cooking instructions, and nutritional information. The dataset itself is not provided publicly, instead, the authors provide a list of image URLs that can be scraped from the Web. While scraping this data, we encountered instances where we could not download images due to an expired URL or a corrupt image. Additionally, some of the recipes in the dataset did not have any accompanying images. Therefore, we only utilized recipes with at least one corresponding image available to us. Following the dataset filtering process, we obtained a training set comprising 259,932 samples, a development set containing 55,773 samples, and a test set consisting of 56,029 samples adequate for recipe analysis. 

\subsection{Baselines}

\noindent \textbf{Ingredient Extraction:}
We present a comparative analysis of \texttt{\textbf{FIRE}}'s ingredient extractor against two retrieval-based techniques: R$_{I2L}$ and R$_{I2LR}$ \cite{s1}. R$_{I2L}$ learns joint embeddings between images and ingredient lists and uses them to retrieve the most relevant recipe within the embedding space. R$_{I2LR}$ expands upon this approach by incorporating the joint embedding between recipe title, instructions, ingredients, and the corresponding food image to further enhance retrieval. We also compare our approach to two state-of-the-art generative models, namely FF$_{TD}$ and InverseCooking  \cite{salvador2019inverse}. FF$_{TD}$ models the joint distribution of the ingredients set by utilizing the target distribution and greedily sampling from a cumulative distribution of sorted output probabilities until the sum of probabilities of selected elements exceeds a specified threshold. InverseCooking  is an attention-based model that takes embeddings from ResNet50 as input and uses a transformer decoder architecture for ingredient generation.


\noindent \textbf{Cooking Instruction Generation:} As baselines, we utilize both InverseCooking \cite{salvador2019inverse} and Chef Transformer \cite{chef-transformer}. We specifically select these two baselines as their code was publicly available, and they work on recipe generation rather than retrieval from a database. 
InverseCooking is an end-to-end recipe generation model that takes food images as input and extracts ingredients, which along with image embeddings, are used to generate the title and the cooking instructions.
Like our method, InverseCooking is also trained on Recipe1M.
Chef Transformer is trained on the RecipeNLG \cite{bien2020recipenlg} dataset and exclusively relies on ingredient inputs rather than food images for cooking instructions generation. Therefore, in the case of the Chef Transformer, we use ground truth ingredients for testing. 

\subsection{Evaluation Metrics}
We evaluate \texttt{\textbf{FIRE}} on end-to-end cooking instruction generation and through ablation study on ingredient extraction.
For end-to-end \textbf{recipe generation}, we employ document-level evaluation metrics: \textit{SacreBLEU} and \textit{RougeL} to assess the quality of our model's output.
Since the output of \textbf{ingredient extraction} models is a set, we evaluate their performance using \textit{F1-score/Dice score}, and \textit{Jaccard/IoU} similarity index, computed for accumulated counts of true positives, false negatives, and false positives over the entire dataset. 

\section{Results \& Analysis} \label{sec:results}

\subsection{End-to-End Recipe Generation}
The results in Table \ref{table-recipe} show that \texttt{\textbf{FIRE}} exhibits superior performance compared to the two SotA baselines, InverseCooking and Chef Transformer. These results demonstrate our proposed pipeline's ability to generate precise and coherent recipes, corroborating the effectiveness of \texttt{\textbf{FIRE}} and emphasizing the value of language generation models for high-quality recipe generation. 
These results also support our expectation that the \texttt{\textbf{FIRE}} method can generalize well without ingredient quantity information given at inference time, even when they were present during training. Meanwhile, we observe that training with extra information results in fewer hallucinations, especially regarding ingredients quantity (e.g., 2 tablespoons of salt) and cooking time (e.g., heat for 10-12 minutes).

\texttt{\textbf{FIRE}} with automatically extracted title and ingredients achieved a relative improvement over InverseCooking of 6\% and 8\% on SacreBLEU and RougeL scores, respectively.
Notably, InverseCooking incorporates both image embeddings and automatically extracted ingredients during the cooking instruction generation phase. Meanwhile, \texttt{\textbf{FIRE}}'s instruction generation language model relies on the recipe title and ingredients only, which provide \texttt{\textbf{FIRE}} with informative signals to generate comprehensive recipes.

As Chef Transformer does not support image input, it uses ground-truth ingredients for cooking instruction generation. In comparison, \texttt{\textbf{FIRE}} faces realistic challenges due to noisy ingredient extraction. Yet, \texttt{\textbf{FIRE}} easily outperforms Chef Transformer, and the gap increases further when \texttt{\textbf{FIRE}} is also provided with a ground-truth title and ingredient set. The low performance of Chef Transformer on this task can be attributed to its architecture and its reliance on just ingredients without any title information. As a set of ingredients can correspond to multiple recipes, the title may be crucial for disambiguation and coherence. For example, both Stir-fried Ginger Chicken and Garlic Ginger Chicken Soup share the same set of ingredients (\textit{chicken, garlic, soy sauce}, and \textit{ginger}). Despite this commonality, this same set of ingredients leads to two entirely different recipes.

\begin{table}[!t]
\centering
\caption{Recipe generation comparison on the test dataset. We report mean with one standard deviation of 10 experiments. \textbf{Bold} represents the best model. ($^+$) represents the model tested on the ground truth title and ingredients to generate the recipe}

\label{table-recipe}
\small
\begin{tabular}{lcccc}
\toprule
 & \textbf{SacreBLEU} & \textbf{ROUGE L }\\
\midrule
Chef Transformer \cite{chef-transformer} & 4.61 $\pm$ 0.32 & 17.54 $\pm$ 0.19 \\
InverseCooking \cite{salvador2019inverse} & 5.48 $\pm$ 0.21 & 19.47 $\pm$ 0.15 \\
\texttt{\textbf{FIRE}} (without \textit{loss}$_{card}$) & 5.91 $\pm$ 0.17 & 20.87 $\pm$ 0.13 \\ 
\texttt{\textbf{FIRE}} (ResNet50) & 5.87 $\pm$ 0.10 & 20.49 $\pm$ 0.08 \\ 
\texttt{\textbf{FIRE}} & 6.02 $\pm$ 0.15 & 21.29 $\pm$ 0.10\\ 
\texttt{\textbf{FIRE}$^+$} & \textbf{7.29} $\pm$ 0.11 & \textbf{25.17} $\pm$ 0.07 \\
\bottomrule
\end{tabular}
\end{table}


\subsection{Ablation Study}

\noindent \textbf{Ingredient Extraction}
The results on the ingredient extraction task are shown in Table \ref{table-ingrs1}.
The retrieval-based approaches (R$_{I2L}$  and R$_{I2LR}$) yield poor results. This can be expected, given their reliance on the presence of an exact matching recipe in the static dataset and their dependence on the dataset size and diversity. The models FF$_{TD}$, InverseCooking, and \texttt{\textbf{FIRE}}, which employ conditional generation, exhibit relatively higher performance in capturing ingredient information from food images. 
Moreover, out of these three models, \texttt{\textbf{FIRE}} achieves the highest IoU and F1 scores among all of the models, surpassing the second-ranked InverseCooking model with a relative margin of 1.5\% in terms of IoU and 1.4\% in terms of F1 score. We attribute this improvement to \texttt{\textbf{FIRE}}'s superior feature extraction capability that uses ViT rather than ResNet50.

\begin{table}[!t]
\centering
\caption{Evaluation results on ingredient extraction using set metrics (IoU and F1). \textbf{Bold} represents the best model.}
\small
\label{table-ingrs1}
\begin{tabular}{lcc}
\toprule
\textbf{Model}  & \textbf{IoU} & \textbf{F1}\\
\midrule
R$_{I2L}$ \cite{s1} &  18.92  & 31.83 \\
R$_{I2LR}$ \cite{s1} &  19.85  & 33.13 \\
FF$_{TD}$ \cite{salvador2019inverse} &  29.82  & 45.94 \\
InverseCooking \cite{salvador2019inverse}  & 32.11 & 48.61 \\
\texttt{\textbf{FIRE}} &  \textbf{32.59}  & \textbf{49.27} \\
\bottomrule
\end{tabular}
\end{table}

\noindent \textbf{Image Feature Extraction}
To understand the observed ingredient extraction gap between \texttt{\textbf{FIRE}} and InverseCooking, we compare the impact of image feature extractors on ingredient extraction. We ablate our feature extractor (ViT) with state-of-the-art CNN models: ResNet18, ResNet50, ResNet101, and InceptionV3.
Table \ref{table-ingrs2} reveals that ViT outperforms the other feature extractors, demonstrating its superior ability to capture and represent food image features relevant to ingredient extraction. 
Furthermore, to assess the feature extractor's influence on the end-to-end FIRE pipeline, we substituted ViT with ResNet50. This change resulted in a performance decrease, as indicated in Table \ref{table-recipe}.
This finding emphasizes the efficacy of leveraging the state-of-the-art feature extractor ViT for improved results in our food computing system.


\begin{table}[t]
    \begin{minipage}[!ht]{0.5\linewidth}
        \centering
        \small
         \caption{Impact of SotA feature extractors on ingredient extraction. All models are trained and tested on 10\% dataset.}
         \small
        \begin{tabular}{lcc}
        \toprule
        \textbf{Model}  & \textbf{IoU} & \textbf{F1}\\
        \midrule
        ResNet18 & 25.88   & 39.31 \\
        ResNet50 &  26.94  & 40.51 \\
        ResNet101 & 26.37   & 40.12 \\
        InceptionV3 & 25.31   & 38.92 \\
        ViT &  \textbf{27.69}  & \textbf{42.73} \\
        \bottomrule
        \end{tabular}
\label{table-ingrs2}
    \end{minipage}
    \hfill
    \begin{minipage}[!ht]{0.45\linewidth}
        \centering
        \small
         \caption{Comparison between zero-shot (ZS) and fine-tuned (FT) versions of BLIP and T5.}
        \begin{tabular}{lcc}
        \toprule
        \textbf{Model}  & \textbf{ZS} & \textbf{FT}\\
        \midrule
        BLIP & 17.89 & 37.72  \\
        T5 & 2.47   & 6.02 \\
        \bottomrule
        \end{tabular}
\label{table-ablation-zero-ft}
    \end{minipage}
\end{table}

\noindent \textbf{Zero-shot vs Fine-tuned} We compare the performance of zero-shot BLIP and T5 model against our fine-tuned model. The outcomes are detailed in Table \ref{table-ablation-zero-ft}. To assess the results for title generation by BLIP, we use a string similarity approach based on the \textit{longest common subsequence} (LCS), as achieving an exact match is infrequent due to the vast array of recipe variations. For example, if the actual title is \textit{`black bean and rice salad'} and BLIP predicts \textit{`black bean and rice,'} then a conventional accuracy metric would yield zero, whereas the LCS score would be 0.76. For T5, we utilize the SacreBLEU metric. The results demonstrates that SotA BLIP and T5 models necessitate task-specific fine-tuning.

\noindent \textbf{Cardinality Loss} 
Complementing $loss_{eos}$ with $loss_{card}$ improves the model's ability to extract the correct ingredients from food images. In contrast, using only binary cross-entropy does not consider dependencies among elements in the set.
We trained \texttt{\textbf{FIRE}}'s ingredient extraction model without cardinality loss (\textit{loss}$_{card}$) to check the impact of adding this loss in model. 
Without cardinality loss, we believe the model struggles in realizing correct number of ingredients, which leads to divergence from the ground truth, thus lowering performance as shown in Table \ref{table-recipe}.

\subsection{Error Analysis} \label{sec:case}
\begin{figure}[!t]
	\centering
\includegraphics[width=\linewidth]{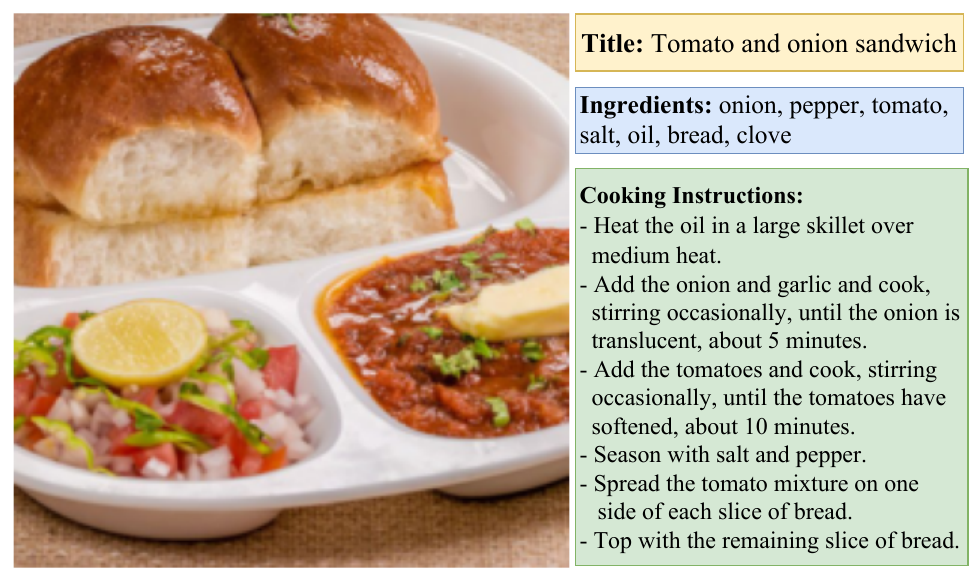}
	\caption{Recipe prediction by \texttt{\textbf{FIRE}} for Pav Bhaji image.}
	\label{fig:paav_bhaji}
\end{figure}

In order to gain further insight into the performance of our recipe generation method, we inspected its performance on individual images. As shown in Figure \ref{t5}, \texttt{\textbf{FIRE}} is often able to generate a correct recipe for dishes similar to those present in the Recipe1M dataset. Meanwhile, we also study its ability to provide a recipe for Pav bhaji, a popular Indian dish that is not present in the Recipe1M dataset.
\texttt{\textbf{FIRE}} generates a recipe for \textit{`tomato onion and sandwich'} as shown in Figure \ref{fig:paav_bhaji}. As expected, the generated recipe is unrelated to the intended dish. Other state-of-the-art models are also not able to predict the correct recipe.
We acknowledge the need for improvement in our model to better generalize to novel recipes.
Meanwhile, we highlight the importance of developing better evaluation metrics. Conventional evaluation metrics such as SacreBLEU and ROUGE, failed to capture the accuracy of the generated recipes and detect certain text hallucinations. Given the significant impact of even a single mistake on the final outcome of a dish, it is crucial to develop a robust metric that can reliably ensure the completion of the desired cooking task beyond text similarity.

\begin{figure*}[!t]
	\centering
	\includegraphics[width=\linewidth]{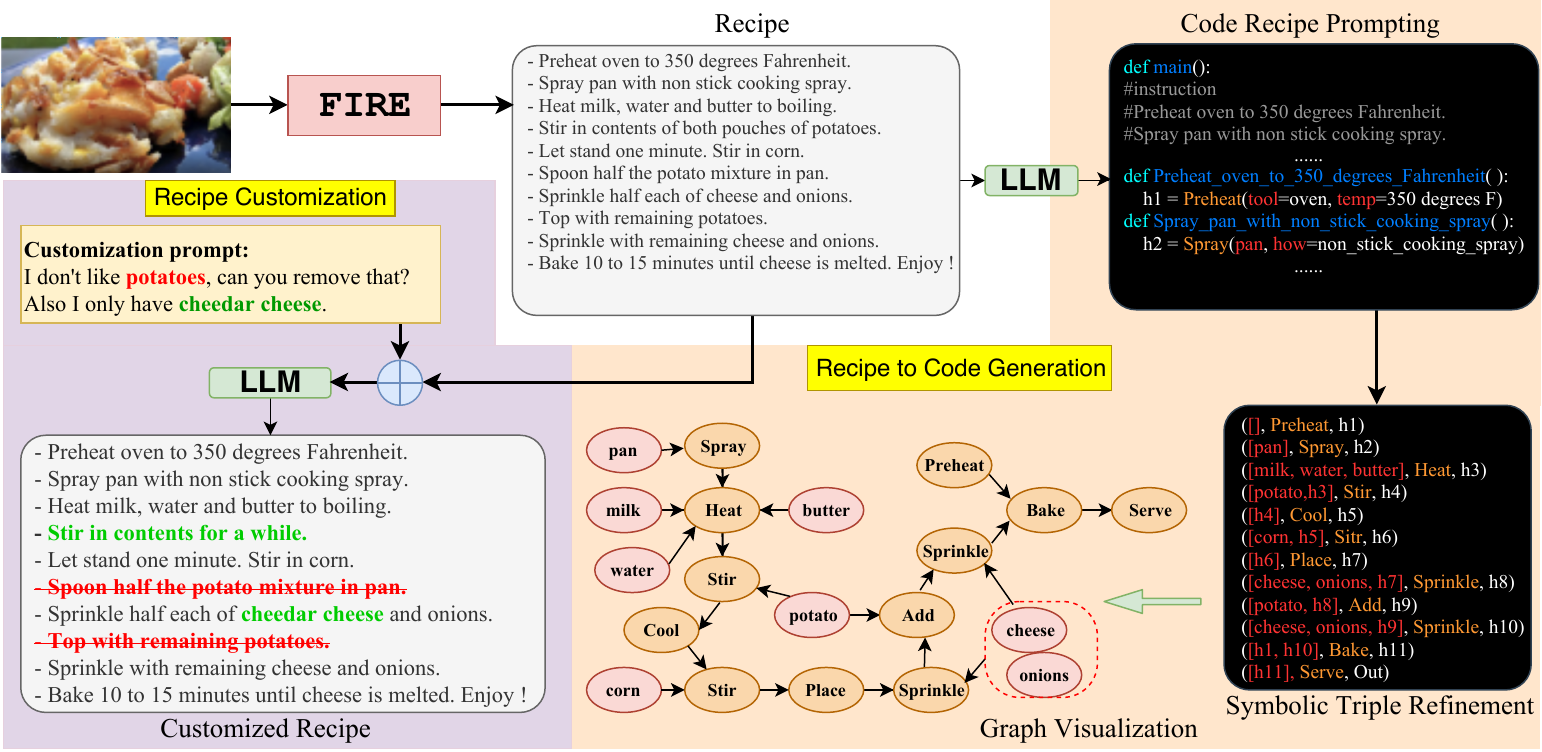}
	\caption{Applications of \texttt{\textbf{FIRE}}: Recipe Customization and Recipe to Code Generation. }
	\label{fig:applications}
\end{figure*}

\section{FIRE Applications} 
While \texttt{\textbf{FIRE}} achieves state-of-the-art performance on the ambitious task of generating recipes from images, we go a step further and investigate its integration into larger pipelines for food computing applications. Namely, considering the promise of few-shot prompting of large language models, we describe how \texttt{\textbf{FIRE}} and large LMs can be integrated to support recipe customization and recipe-to-machine-code generation. For both applications, we provide a potential pipeline with an illustrated recipe and we conduct a pilot study to investigate its potential. 

\label{sec:applications}
\subsection{Recipe Customization}
Recipe customization is crucial due to the connection between food, customs, and individual preferences. Additionally, it becomes essential when addressing allergies or dietary restrictions.
Surprisingly, despite the evident demand, existing literature lacks dedicated efforts in the domain of recipe customization. 
We are inspired by the Computer Cooking Contest (CCC) \cite{ccc}, an annual event showcasing computational systems that generate novel and creative recipes, enabling participants to employ AI and computational creativity in exploring innovative food combinations and techniques.
However, we cannot use CCC directly because judges perform its evaluation manually.
Our work aims to bridge the research gap by enabling personalized recipe customization, considering individual taste profiles and dietary restrictions. 
To guide future research in this area, we showcase the ability of \texttt{\textbf{FIRE}} to support a recipe customization approach that focuses on a wide range of topics 
 (e.g. ingredient replacement, taste adjustment, calories adjustment, cooking time adaptation) to test few-shot performance thoroughly.
As shown in the purple part of Figure \ref{fig:applications}, we perform \textit{ingredient removal} to trim the potatoes from the recipe. Two sentences related to potatoes are deleted in the modified version, and one sentence is modified to ensure consistency. Specifically, we perform \textit{ingredient addition} to replace `cheese' with `cheddar cheese' and recognize that it should be added before baking, resulting in the modified sentence `Sprinkle half each of cheddar cheese and onions.'
We manually design four demonstrations to hint GPT-3 to solve the customization requirements. 

\noindent \textbf{Analysis} 
In order to assess the effectiveness of recipe customization, we conducted a human evaluation with seven experts involving 10 recipes and their customizations. Evaluators rated four attributes: efficacy, coherence, soundness, and proportions and measurements, on a 0 to 4 scale (0: strongly disagree, 1: disagree, 2: neutral, 3: agree, 4: strongly agree). Table~\ref{tab:rc} shows the average experimental results. On average, each attribute has a high result (3.5 to 3.76) with high Fleiss~\cite{fleiss1971measuring} kappa inter-annotator agreement (0.78 to 0.92). The results indicate the promise of integrating our method with few-shot LM prompting. Albeit provided with a limited number of demonstrations, the model can handle complex examples like \textit{Can you make the food with fewer calories?} and replace \textit{milk} with \textit{almond milk}. 
We refer the reader to the Appendix for further details about the design, the data, and the results of the pilot experiment.




\subsection{Generating Machine Code for Image-based Recipes} 
Converting recipes to machine code enables automation, scalability, and integration with various systems, reducing manual intervention, resulting in savings in labor costs and reducing human errors while preparing the food. To facilitate this task, we combine \texttt{\textbf{FIRE}}'s recipe generation strength with the ability of large LMs to manipulate code-style prompts for structural tasks~\cite{madaan22emnlp}. 
We show an example approach for generating Python-style code representations of recipes generated by \texttt{\textbf{FIRE}}, by prompting GPT-3 (orange part in Figure \ref{fig:applications}). 
This envisioned approach has two phases: code recipe prompting and symbolic triple refinement.

\noindent \textbf{Code Recipe Prompting} We convert the output of \texttt{\textbf{FIRE}} into a Python-style prompt and leverage GPT-3 to generate code representations as shown in Figure \ref{fig:applications}. Further, we refine these representations into symbolic triple representations within a predefined space. For all input recipes $r \in R$, we constructed corresponding Python-style prompting $r^p$ and code representation $c$. For any new recipes $r^\prime$, the input to the prompting pipeline was $r^p_1 \oplus c_1 \oplus \cdots \oplus r^p_k \oplus c_k \oplus r^\prime$, where $k=4$ was the number of demonstrations and the output code representation $c^\prime$ is completion result of GPT-3. 

\noindent \textbf{Symbolic Triple Refinement} For further use in industrial applications, we refine code generation into symbolic triples $(i, r, o)$, where $i$ and $o$ represent the input list and output of operations, and $r$ represented the cooking instruction and parameter details. 
This allows for a more structured and standardized representation of the generated code, facilitating easier integration with various applications.

\noindent \textbf{Analysis} We conduct a similar human evaluation process focusing on how well ingredients, cooking instructions, and their descriptions are translated to code format on a scale of 0 (extremely poor) to 5 (excellent). Each property is rated on average between 4.27 and 4.47, with an inter-annotator agreement between 0.75 and 0.83. Despite the promising experiment results, few-shot prompting can produce hallucinations when tracking ingredients, especially in long contexts or when similar cooking tools are involved (e.g., saucepan, frying pan), which can be further explored by future work. We refer the reader to the Appendix for details about the design, the data, and the results of the pilot study.




\section{Conclusion \& Future Work} \label{sec:conclusion}
This paper introduced \texttt{\textbf{FIRE}}, a methodology tailored to the field of food computing, focusing on generating the food title, extracting ingredients, and generating cooking instructions solely from image inputs. We leveraged recent advancements in CV and language modelling to achieve superior performance against strong baselines. Furthermore, we demonstrated the practical applications of \texttt{\textbf{FIRE}} for recipe customization and recipe-to-code generation, showcasing the adaptability and automation potential of our approach. Experimental results validated the efficacy of \texttt{\textbf{FIRE}}, highlighting its promising prospects for future advancements and wide-ranging adoption in food computing. Inspired by our experiments, we list three challenges that should be addressed in future research:  
\begin{enumerate} 
\itemsep0em
    \item  A major limitation of both the proposed work and existing approaches lies in the absence of a reliable grounding mechanism \cite{book-grounding} to ascertain the correctness of generated recipes. Conventional metrics are insufficient to capture this challenge. 
    We propose to address this limitation by developing a metric that effectively captures the coherence and plausibility of generated recipes, providing a more comprehensive evaluation framework for recipe generation systems.
    \item The diversity and availability of recipes are heavily dependent on the locations, climates, and religions~\cite{min2017you,zhu2013geography,simas2017food}, which prevent users from preparing food based on predefined recipes. One solution can be the injection of knowledge graphs~\cite{ilievski2021cskg,Sap2019ATOMICAA}, which reflect the connection between the ingredients based on symbolic relations and contextual factors, thus informing the models about alternative ingredients.
    \item Hallucination remains a critical challenge in recipe generation by natural language and vision models. We will investigate the possibility of incorporating methods for state tracking of participants\cite{ma2022coalescing,jiang2023transferring} to enhance the production of reasonable and accurate results. 
\end{enumerate}

{\small
\bibliographystyle{ieee_fullname}
\bibliography{egbib}
}

\clearpage

\onecolumn

\noindent \textbf{\large APPENDIX}

\appendix


\begin{appendices}
\renewcommand{\thesection}{A\arabic{section}}

\section{Hyper-parameters}

\noindent \textbf{Title Generation Model:} For title generation, we utilized the BLIP Salesforce/blip-image-captioning-base\footnote{\url{https://huggingface.co/Salesforce/blip-image-captioning-base}} model, which was trained for 20 epochs with a batch size of 24 and a learning rate of $10^{-5}$. To evaluate the performance on the validation set, we opted to employ string similarity metrics, specifically the longest common subsequence (LCS), instead of traditional loss functions. This decision was motivated by the superior results achieved by the LCS-based model in comparison to loss based validation function.

\noindent \textbf{Ingredients Generation Model:} We train the model for 100 epochs with a batch size of 150, utilizing an Adam optimizer with a learning rate of $10^{-4}$. At each epoch, we decrease the learning rate by 0.01\%. The input image was resized to a dimension of 224$\times$224$\times$3 (299$\times$299$\times$3 for InceptionV3), while the ingredients vocabulary size was set to 1488, with a corresponding embedding size of 512.

\noindent \textbf{Recipe Generation Model:} We fine-tune the base T5 (220M parameters) with a batch size of 12 and learning rate of $3\times10^{-4}$ for 30 epochs using an AdamW optimizer. We have \texttt{max source length=50} and \texttt{max target length=512} considering the average length of  titles, ingredients, and cooking instructions. During generation, we use beam search with \texttt{num beams=4}, \texttt{length penalty=1}, and \texttt{repetition penalty=2.5}.

\begin{table*}[!h]
    \caption{Results of human evaluation on recipe customization}
    \label{tab:rc}
    \centering
    \begin{tabular}{|l|c|c|c|c|}
    \hline
    Metric& Efficacy&Coherence&Soundness&Proportions and Measurements\\
    \hline
        Average Score&3.59 & 3.76 & 3.76 & 3.54 \\ \hline
        Inter-annotator agreement&0.90 & 0.92 & 0.89 & 0.78 \\ \hline
    \end{tabular}
\end{table*}

\begin{table*}[!h]
    \caption{Results of human evaluation on recipe to code generation.}
    \label{tab:cg}
    \centering
    \begin{tabular}{|l|c|c|c|}
    \hline
    Metric & Ingredients & Cooking instructions & Step parameter\\
    \hline
        Average Score&4.47& 4.29 & 4.27 \\ \hline
        Inter-annotator agreement&0.83 & 0.75 & 0.79 \\ \hline
    \end{tabular}
\end{table*}

\section{Resource Details}
We used a hardware configuration consisting of four Quadro RTX 8000 GPUs, each equipped with 48 GB GDDR6 memory, to train the Image to Title and Ingredients model. The GPU computations were supported by an Intel(R) Xeon(R) Gold 5217 CPU running at 3.00GHz, which offered 32 processors. To train the Cooking Instructions model, we employed eight NVIDIA RTX A5000 GPUs, with each GPU having 24 GB GDDR6 memory. The GPU acceleration was complemented by an Intel (R) Xeon(R) Gold 5215 CPU operating at 2.50GHz, featuring 40 processors.

\section{Dataset}
\noindent As of June 2023, the Recipe1M dataset is not publicly available. However, we accessed the data using the following URL: \url{http://im2recipe.csail.mit.edu/dataset/login/}.

\section{Pilot Experiment}

\subsection{Recipe Customization}
Table~\ref{tab:rc} shows the detailed experimental results on the average over seven human annotators. We utilized a scaling system ranging from 0 to 4 to express the level of agreement, where 0: strongly disagree, 1: disagree, 2: neutral, 3: agree, and 4: strongly agree. Four evaluation criteria were chosen to assess the customized recipe.
\begin{enumerate}
    \item \textbf{Efficacy}: The customization is implemented correctly, and the modified recipe successfully achieves the desired taste or flavor profile based on the prompt.
    \item \textbf{Coherence} The modified recipe maintains a coherent structure and flow, ensuring that the added or altered ingredients make sense within the context of the original recipe.
    \item \textbf{Soundness} The instructions are clear and unambiguous. A user can easily follow the steps to recreate the customized recipe.
    \item \textbf{Proportions and Measurements} The proportions and measurements of the modified ingredients are appropriate and maintain a balance with the original recipe.
\end{enumerate}

\noindent \textbf{Survey Link:}\\ \url{https://bit.ly/fire-customization}

\subsection{Recipe to Code Generation}
We conducted a survey to assess the quality of the code generated from recipes. Each instance in the survey was evaluated based on three questions. 
Firstly, we determined whether all ingredients were correctly converted to code. Secondly, we examined the accuracy of converting cooking instructions to code. Lastly, we assessed whether cooking instruction parameters were appropriately converted.  
The questions asked are as follows.
\begin{enumerate}
    \item \textbf{Ingredients:} How well are the ingredients incorporated into the code?
    \item \textbf{Cooking instructions:} How well are the cooking instructions translated into code?
    \item \textbf{Step parameter:} How well are the step parameters of cooking instructions translated into code?
\end{enumerate}
Participants were asked to rate each question on a scale from 0 to 5, 
where 
\textbf{0: Extremely poor} - The conversion of ingredients is completely inadequate.
\textbf{1: Low quality} - Only a few ingredients are converted correctly, and most are missing.
\textbf{2: Fair quality} - Some ingredients are accurately incorporated, but there are instances of misleading information or unclear details.
\textbf{3: Good quality} - A significant portion of the ingredients are accurately incorporated.
\textbf{4: Very good quality} - All ingredients are mostly incorporated correctly, with only minor errors.
\textbf{5: Excellent quality} - Every single ingredient is flawlessly incorporated, ensuring accurate representation. Table~\ref{tab:cg} shows the detailed experiment results on the average over seven human annotators.

\noindent \textbf{Survey Link:}\\ \url{https://www.bit.ly/fire-code-generation} \\

Moreover, we conduct a pilot experiment on a sample of 200 recipes. To describe cooking details, we predefine 30 cooking operation functions, such as heat, cut, and boil, with eight function parameters, including time and tools. The defined set covers 84\% of operations that appear in the dataset.

\section{Prompting Design for Downstream Applications}

\subsection{Recipe Customization}
Our test on recipe customization covers a wide range of topics, following is the list of prompts used in each topic:

\begin{enumerate}
    \item \textbf{Ingredient adjustment}: I (don't) like \{ingredient\}, can you remove/add that for me? 
    \item \textbf{Detail addition}: I am new to cooking, can you expand more details for the recipe?
    \item \textbf{Taste adjustment} I (don't) like \{taste\} food, can you change the taste of the recipe?
    \item \textbf{Calories adjustment} Can you reduce/increase the calorie content of the food?
    \item \textbf{Cooking time adaptation} Can you provide a more convenient version that can be done in \{time\} minutes?
\end{enumerate}

\subsection{Recipe to Code Generation}
\noindent For the conversion of the recipe to Python code, we formulate explicit rules for the demonstration and employ well-defined prompts to provide clear guidance to the LLMs. For each recipe `r' with a list of cooking instructions \{$c_1$,$c_2$,$\dots$,$c_n$\}, its code format conform following format:
\\
\begin{quote}
\hspace*{2cm}def main():\\
\hspace*{3cm}\#instruction\\
\hspace*{3cm}\#$c_1$\\
\hspace*{3cm}\#$c_2$\\
\hspace*{3cm}$\dots$\\
\hspace*{3cm}\#$c_n$\\

\hspace*{3cm}def $f_1$():\\
\hspace*{4cm}$h_1$ = \{operation\}(input, (tool),(how),(temperature))\\
\hspace*{3cm}def $f_2$():\\
\hspace*{4cm}$h_2$ = \{operation\}(input, (tool),(how),(temperature))\\
\hspace*{3cm}$\dots$\\
\hspace*{3cm}def $f_n$():\\
\hspace*{4cm}$h_n$ = \{operation\}(input, (tool),(how),(temperature))\\
\end{quote}

The code script begins with comments on cooking instructions. Each cooking instruction `c' will become function name `f' in the code script by adding an underscore between each word within cooking instructions. For example, ``\textit{Preheat oven to 350 degrees Fahrenheit.}" will become \\$Preheat\_oven\_to\_350\_degrees\_Fahrenheit.$ For each function definition, we will extract the operation as the function object and input ingredients as parameters. The parameter set will also consider tools, cooking methods, and temperature if they are adaptable. For the same example, ``\textit{Preheat oven to 350 degrees Fahrenheit.}", the function object in the definition will be $h_1$ = Preheat(tool = oven, temp = 350 degrees F).

\end{appendices}

\end{document}